\documentclass[runningheads]{llncs}

\usepackage{eccv}

\usepackage{eccvabbrv}

\usepackage{graphicx}
\usepackage{booktabs}

\usepackage{url}
\usepackage{multirow}
\usepackage{multicol}
\usepackage{caption}
\usepackage{array} 
\usepackage{graphicx}
\usepackage{makecell}
\usepackage[accsupp]{axessibility}  
\usepackage{wrapfig}

\usepackage{hyperref}

\usepackage{orcidlink}

\begin{document}

\title{Towards Error-Free Long Video Generation} 

\titlerunning{Abbreviated paper title}

\author{Shuning Chang\textsuperscript{\rm 1}\textsuperscript{\rm 2}\quad Weihua Chen\textsuperscript{\rm 2} \quad Jiasheng Tang\textsuperscript{\rm 2} \quad Hao Xu\textsuperscript{\rm 2} \\ Zeyu Zhang\textsuperscript{\rm 1}\textsuperscript{\rm 2}
Hangjie Yuan\textsuperscript{\rm 2}\quad Yu Lu\textsuperscript{\rm 1} \quad Ruigang Niu\textsuperscript{\rm 2}\quad Fan Wang\textsuperscript{\rm 2} \\ Bohan Zhuang\textsuperscript{\rm 1}\textsuperscript{\rm 2} \quad Yi Yang\textsuperscript{\rm 1}
}

\institute{Zhejiang University \and Alibaba Group}

\maketitle

\begin{abstract}
  Recent advances in video generation have made minute-level synthesis possible; however, generating long videos remains challenging due to error accumulation, attribute drift, and the limited availability of long video data. In this paper, we introduce an infinite-length video generation framework that focusing on addressing these issues and produces high-quality, dynamic, and identity-consistent single-shot long videos. We first finetune a diffusion model as a video extension model on large-scale short video data to autoregressively generate temporally coherent clips. Inspired by the success of large language models (LLMs), we adopt causal attention computation between clips to further finetune this model on long video data. In this way, the tokens in one clip (short video) are computed by bidirectional attention while tokens among clips are computed by unidirectional attention. This design leverages the strengths of modern diffusion models while preserving long-term context information, effectively mitigating error accumulation and attribute drift. To achieve memory efficiency during inference, we adopt a key-value (KV) caching mechanism to maintain a constant KV memory. Furthermore, we introduce truncation-rectified flow (T-RFlow) technique to further suppress error accumulation. Experimental results demonstrate the effectiveness of our method. 
  Our framework establishes a new benchmark for realistic and coherent minute-level video synthesis.
\end{abstract}

\section{Introduction}
\label{sec:intro}

\begin{figure}[t]
\begin{subfigure}{0.49\textwidth}
\includegraphics[width=1.0\linewidth]{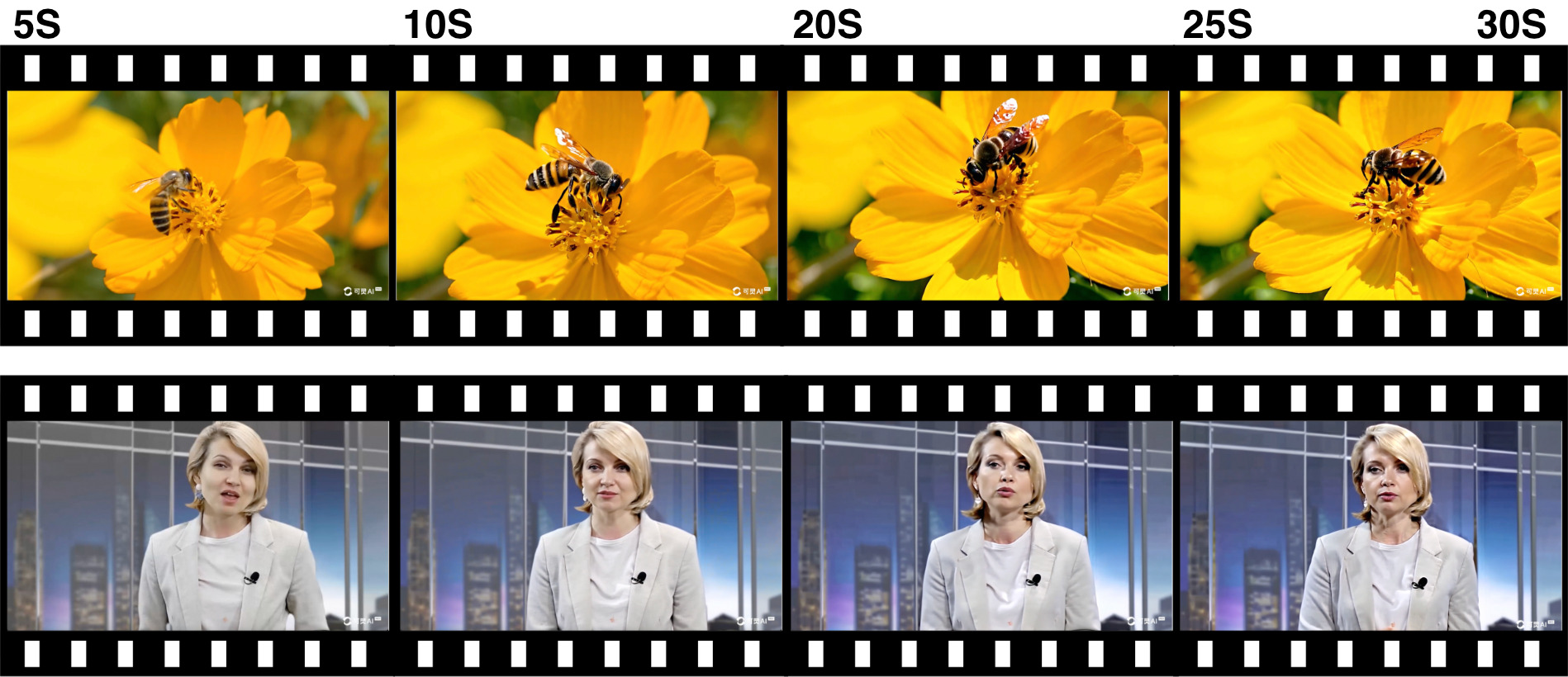}
\centering
\caption{Long videos generated by the commercial-grade model Kling exhibit error accumulation, including excessive sharpness and color saturation.}
\label{figure1a}
\end{subfigure}
\begin{subfigure}{0.49\textwidth}
\includegraphics[width=1.0\linewidth]{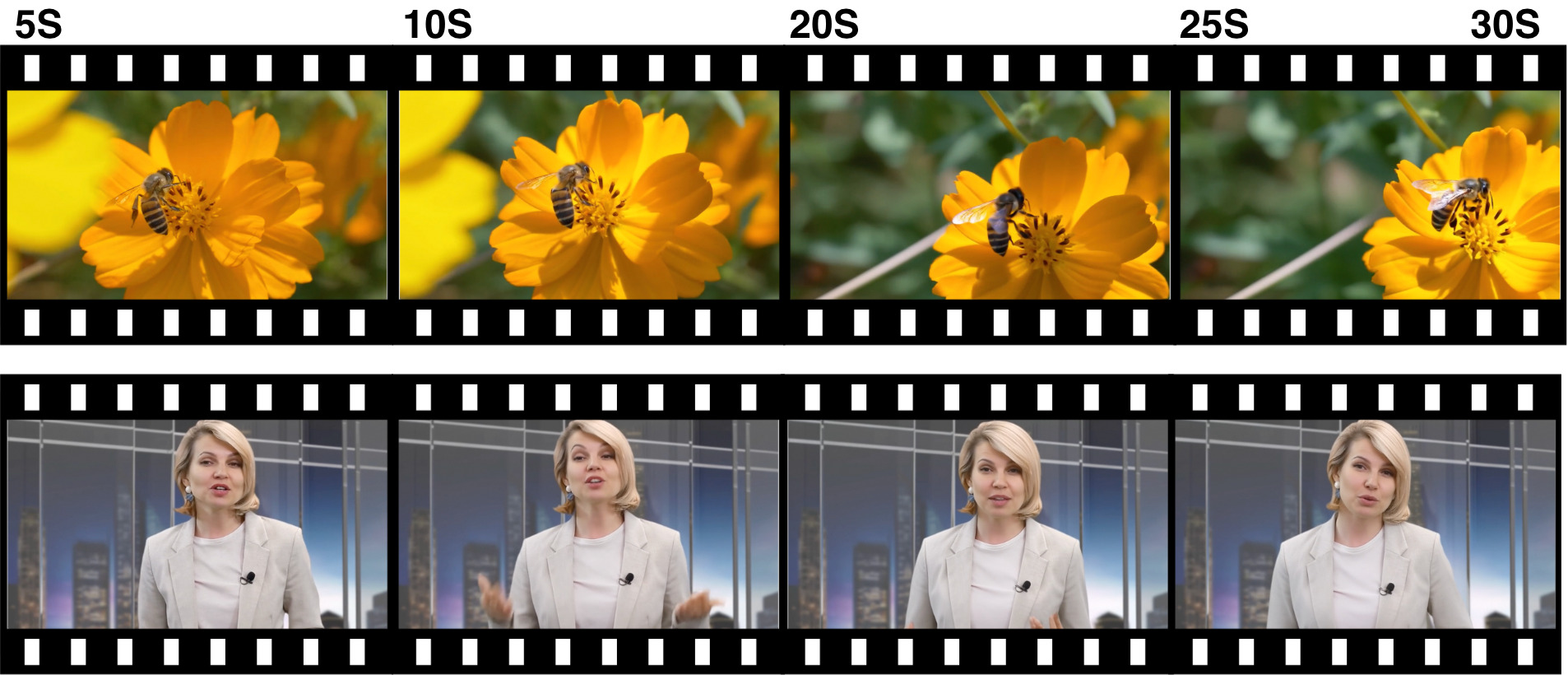}
\centering
\caption{Long videos produced by our model demonstrate temporal coherence, effectively mitigating error accumulation.}
\label{figure1b}
\end{subfigure}
\centering
\caption{30-second videos generated by both the Kling model and our model using the same initial frames.\vspace{-3mm}}
\label{figure1}
\end{figure}

Long video generation has gained increasing attention as it aims to surpass the duration limitations of current video foundation models~\cite{wan2025,kong2024hunyuanvideo,yang2024cogvideox,opensora2,genmo2024mochi,lin2024open}, which typically produce only 5–10 second clips, and extend generation to minute-level or even unlimited lengths. Existing approaches can be broadly categorized into two paradigms: (1) fine-tuning video foundation models~\cite{voleti2022mcvd,harvey2022flexible,yang2023video,he2022latent} into video extension models, and (2) adopting autoregressive (AR) generation frameworks~\cite{gu2025long,wang2024loong}. The former typically lacks long-term memory, while the latter abandons the advantages of recent diffusion-based video generation models~\cite{wan2025,kong2024hunyuanvideo}, which have shown strong performance in short video synthesis. Error accumulation and attribute drift are the two most critical problems in long video generation, especially single-shot long video generation. Error accumulation refers to the iterative degradation visual quality over time, whereas attribute drift refers to the fading of memory as the model struggles to remember earlier content and maintain temporal consistency. As shown in Figure~\ref{figure1a}, even commercial-grade model like Kling~\cite{Kling} exhibit noticeable degradation over time, such as exposure bias and excessive sharpness. Current AR-based approaches can alleviate below issues but still suffer from low visual quality. Moreover, the shortage of long video data also hampers the progress of this field.

To address these limitations, we begin by analyzing the root causes of error accumulation and attribute drift. We reformulate the rectified flow sampling process---commonly used in advanced diffusion models---as a linear combination of predictions in all the timesteps. We observe that video generation evolves from low-frequency to high-frequency structures, and that error accumulation tends to arise in the low- and high-frequency phases, manifesting as color bias, exposure bias and over-sharpening, while the mid-frequency phase generally remains clean and stable. Since long-form video generation requires conditioning on previous clips to maintain temporal continuity, these errors are compounded over time. Moreover, attribute drift primarily stems from excessive reliance on local information and a lack of long-term context.

According to above observation, we propose a new long video generation framework. Our approach combines the strengths of both diffusion models and autoregressive models. 
We adopt the state-of-the-art diffusion-based model as our pretrained model. We apply bidirectional attention within each clip to preserve the capabilities of original model, while using unidirectional attention across clips to incorporate long-range temporal dependencies in an autoregressive manner.
Specially, the training process is divided into two stages. First, we fine-tune the base diffusion model as a video extension model on large-scale short video datasets, enabling it to generate temporally coherent clips conditioned on the last frames of the preceding clip (conditional frames). This allows the model to learn local continuity. Second, we further fine-tune the model on long video data, computing the KV representations of past clips and feeding them into the current generation step. During inference, we employ the conditional frames and KV from past clips to generate current clip and cache current KV for next-clip generation. Therefore, our method adopts full attention mechanism in each clip to avoid additional inductive bias and causal attention among clips to leverage efficient KV cache. This makes the model aware of long-term dependencies, thereby alleviating both error accumulation and attribute drift.
Unlike large language models (LLMs), which can afford full-context KV memory, video generation normally incurs significantly higher memory costs. To address this, we propose a clip selection mechanism based on cross-clip relevance. We offline compute the semantic correlation between each prompt and its historical clips using an LLM and retain only the KVs of the most relevant clips. This strategy achieves a favorable balance between long-range dependency modeling and memory efficiency.

To further suppress error accumulation in the autoregressive generation, we introduce truncation-rectified flow (T-RFlow). Our T-RFlow decomposes the rectified flow into different frequency components; we remove high-frequency components to mitigate over-sharpening and discard low-frequency elements to reduce color bias while compensating for high-frequency information rate. Benefiting from KV cache and truncation-rectified flow, our method can generate nearly infinite-length high-quality and consistent video as shown in Figure~\ref{figure1b}.


The contributions of this paper are summarized as below.
\begin{itemize}
	\item We propose a novel long video generation framework. We combine diffusion model and AR model, which uses unidirectional attention in each clip and uses didirection attention among clips. Our model incorporate local information to generate continuous frames and long-term context information to generate coherent content, effectively alleviate error accumulation and attribute drift.
	\item To further counter error accumulation. We analyze the reasons of error accumulation, and present truncation rectified flow and inverse technique.
        \item Extensive experiments demonstrate the effectiveness of our method. Our method achieves significant VBench scores on long-video settings.
\end{itemize}

\section{Related work}

Unlike video foundation models~\cite{bar2024lumiere,chen2023videocrafter1,chen2024videocrafter2,guo2023animatediff,hong2022cogvideo,kong2024hunyuanvideo,zhang2025show,chang2026sparsedit} that are generally restricted to only a few seconds, minute-long video generation can be categorized into three main settings: single-shot video generation, multi-shot video generation, and movie-style video composition.

Single-shot generation aims to produce a minute-long segment within a consistent scene and semantic context, prioritizing long-range temporal coherence and visual stability. Some approaches~\cite{voleti2022mcvd,harvey2022flexible,yang2023video,he2022latent,chen2023seine} achieve this by fine-tuning video foundation models into video extension models for extended video generation. Others employ autoregressive (AR) and semi-autoregressive (semi-AR) strategies. AR methods, such as FAR~\cite{gu2025long} and Loong~\cite{wang2024loong}, conceptualize long video generation as a process of next-frame (or next-segment) prediction. Semi-AR methods, on the other hand, generate videos in chunks, executing iterative diffusion-based denoising within each chunk~\cite{peebles2023scalable,yin2023nuwa,hong2024slowfast,yan2025long,zhang2025blockvid,yang2025longlive}. A critical design element in these methods is chunk-level causal conditioning: MAGI-1~\cite{teng2025magi}, Skyreel-V2~\cite{chen2025skyreels}, and Self Forcing~\cite{huang2025self} proceed strictly sequentially across chunks, while FramePack~\cite{zhang2025packing} uses a symmetric schedule that takes guidance from both ends and fills the middle autoregressively. In practice, semi-AR methods typically depend on careful KV cache usage for efficiency and stability over extended durations.

Multi-shot generation generally focuses on managing camera movements and transitions across different scenes or semantic contexts. Recent systems, like LCT~\cite{guo2025long}, RIFLEx~\cite{zhao2025riflex}, and MoC~\cite{cai2025mixture}, organize text–video units with interleaved layouts and positional extrapolation to effectively handle multiple shots.

Movie-style generation aims to create cinematic content by piecing together multiple chunks, often involving various scenes and styles, while maintaining a coherent global narrative or theme. Methods such as VideoTTT~\cite{dalal2025one}, MovieDreamer~\cite{zhao2024moviedreamer}, MovieBench~\cite{wu2025moviebench}, and Captain Cinema~\cite{xiao2025captain} resemble film editing, blending diverse shots into a single cohesive video guided by chunk-level text descriptions.

In this work, we primarily address the challenge of error accumulation over time in long video generation. Since multi-shot and movie-style generation approaches can restart shots, they are less affected by error accumulation issues. Therefore, our focus is on single-shot long video generation.

\section{Method}
\subsection{Preliminaries and problem formulation}
\paragraph{Diffusion models.}
Diffusion Models~\cite{ho2020denoising} are generative models which target distribution $x_0\sim q(x)$ by learning a denoising process with arbitrary noise levels. To this end, a diffusion process is defined to gradually corrupt $x_0$ with Gaussian noise. To reduce computational demands, latent diffusion models (LDMs)~\cite{rombach2022high} propose to modeling the diffusion-denoising process in latent space instead of raw pixel space. This is achieved by using a pretrained variational autoencoder (VAE) $\mathcal{E}$ to compress $x_0$ into a lower dimensional latent representation $z_0=\mathcal{E}(x_0)$. The model is trained with Rectified Flow (RF) framework~\cite{liu2022flow,lipman2022flow,esser2024scaling}, where the noisy sample is a linear interpolation between clean data $z_0$ and sampled Gaussian noise $\epsilon$, \ie, $z_t=(1-t)z_0+t\epsilon$. The training objective is regress the velocity field, $\mathcal{L} =\mathbb{E}_{t, x_0, x_1, c}||u(z_t, t, c)-(\epsilon-z_0)||_2^2$, where $t\in[0,1]$ is the diffusion timestep, $u_{\theta}(\cdot)$ is the neural network, $c$ is the text prompt condition.
\paragraph{Problem formulation.}
In this paper, we aim to achieve autoregressive long-form video generation based on latent video diffusion models (VDMs). Let $V=[v^0, v^1, \cdot\cdot\cdot,v^M]$ be the long video composed of $M$ clips, and each clip $v^i$ has their individual prompt $c_i$. Let $z_0^{1:M}=[z_0^1,\cdot\cdot\cdot,z_0^M]\in\mathbb{R}^{N\times H\times W \times C}$ be the latent sequence encoded by a pretrained VAE, where $N$ is the number of frames of the long video, $H\times W$ is the downsampled resolution, and $C$ is the number of channels. Our goal is to learn a video diffusion model on the latent space: $p_{\theta}(z_0^{1:M}|c_{1:M})$, where c denotes text conditions, and $\theta$ is implemented by a denoising network, modeled as $u_{\theta}(z_t^{1:M}, c, t)$. In the inference phase, we generate the video clip by clip autoregrssively. Each autoregrssion step consists $T$ denoising steps, in which each denoising step samples $z_{t-1}^i\sim p_{\theta}(z_{t-1}^i|z_t^{i}, z_0^{1:i}, c_{1:i})$, \ie, conditioned on the text prompts and previous generated clips.

\begin{figure*}[t]
        \centering
        \includegraphics[width=1.\linewidth]{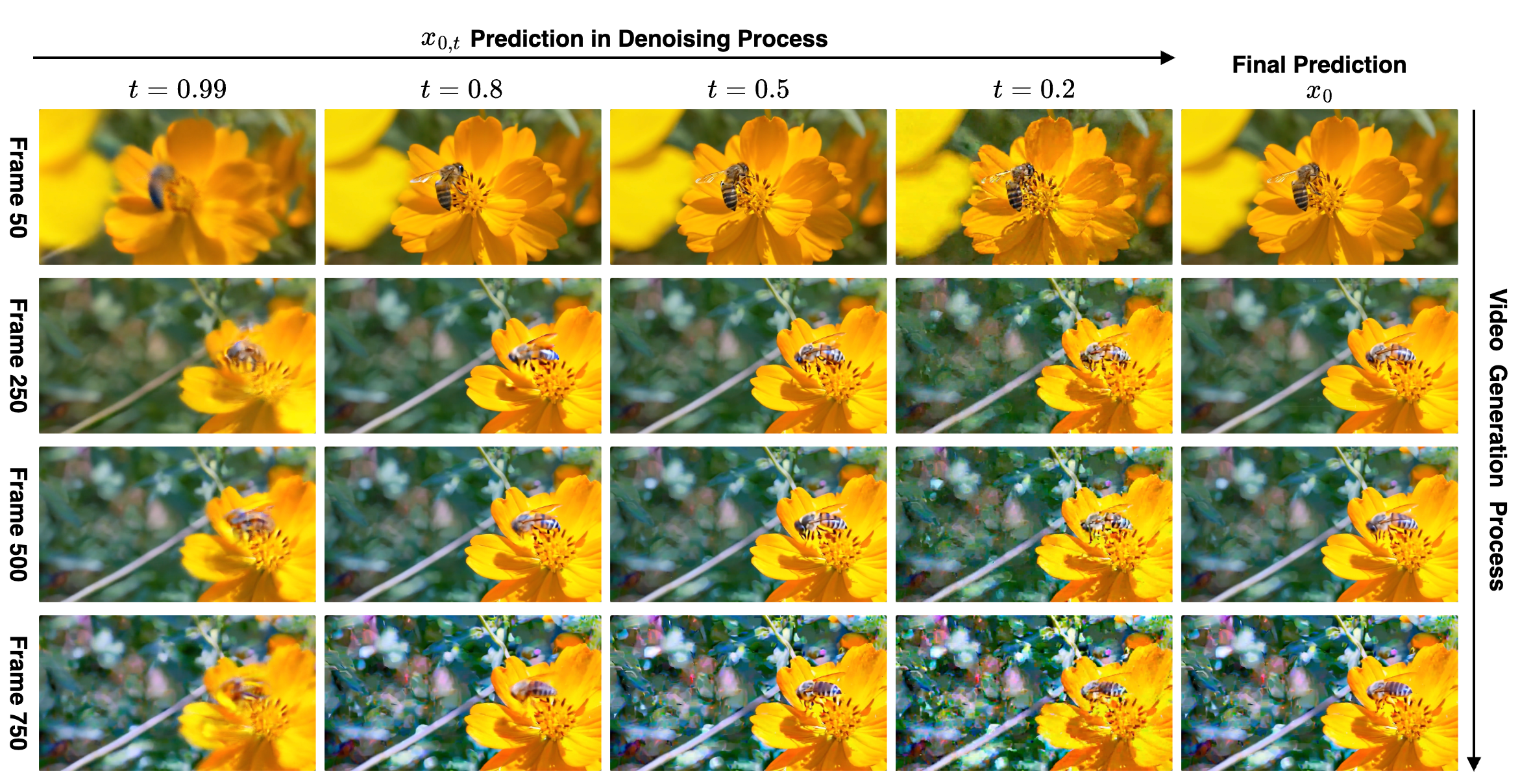}
\caption{Visualization of error accumulation trend along denoising process and video generation process. In the dimension of denoising process, the error accumulation, such as oversaturation, exists in the late timesteps. In the dimension of video generation process, the errors are irreversibly amplified over time. \vspace{-2mm}}
\label{fig:error}
\end{figure*}

\subsection{Analysis of error accumulation and attribute drift}
\label{sec3.1}
To gain a deeper understanding of the sources of degradation in long-form video generation, we introduce a new perspective of the rectified flow (RF) denoising process. 
This formulation allows us to explore how errors accumulate across timesteps by decomposing the output of RF.

RF forecasts a linear trajectory from 
$\epsilon$ to $z_0$. The denoising process over $T$ timesteps is expressed as:
\begin{equation}
\label{eq1}
    \epsilon -\sum_{i=1}^T d_iv_i = z_0,
\end{equation}
where $z_0$ is the predicted clean latent, and $d_i$ is the timestep interval between $t_{i+1}$ and $t_i$, \ie, $di = t_{i+1}-t_{i}$, subject to $\sum_{i=1}^T d_i = 1$. The term $\epsilon$ can be further expressed as a combinations of t items weighted by $d_i$, \ie, $\epsilon=\sum_{i=1}^T d_i\epsilon_i$. Substituting this into Eq~\ref{eq1} becomes:
\begin{equation}
\label{eq2}
    \sum_{i=1}^T d_i(\epsilon -v_i) = z_0.
\end{equation}
Denoted by $v:=\epsilon-z_0$, $\epsilon-v_i$ can be regarded as the estimation of $z_0$ in timestep $i$, represented by ${z_{0,i}}$. Therefore, Eq~\ref{eq2} can be written as:
\begin{equation}
\label{eq3}
    \sum_{i=1}^T d_i{z_{0,i}} = z_0.
\end{equation}
This formulation reveals a new perspective: the final result $z_0$ can be interpreted as an ensemble of the timestep-wise estimation of $z_0$,  and their influence strengths are determined by the weight $d_i$.
This perspective allows to independently observe and operate each timestep's contribution to the overall prediction, facilitating the tracking of error accumulation throughout the denoising process.

We adopt a video extension model based on WanX-2.1~\cite{wan2025}. During inference, We sample some frames and decompose them into timestep-wise latent $z_0^i$. We use a VAE decoder $\mathcal{D}$ to reconstruct the videos $x_{0,i}$ from the latent $z_{0,i}$. We visualize the result in Figure~\ref{fig:error}, which allows us to analyze error accumulation from two perspectives: the denoising process and the video generation process. From the denoising perspective, we observe that the rectified flow (RF) process proceeds from low-frequency to high-frequency reconstruction. As illustrated in each row of Figure~\ref{fig:error}, most visual degradation originates from the high-frequency components introduced during the late timesteps (e.g., $t=0.2$). In contrast, the middle timesteps (e.g., $t=0.8$ and $t=0.5$) produce more stable and accurate predictions. We argue that current diffusion foundation models~\cite{wan2025,Kling,kong2024hunyuanvideo}, having been extensively trained with large-scale high-quality data and substantial GPU resources, possess strong modeling capacity for real-world distributions. The middle timesteps are normally sufficient to reconstruct the majority of visual content, while the later timesteps tend to focus on high-frequency signals—including imperceptible real-world noise. While such noise may be negligible in short video generation, it can be significantly amplified in the context of long video generation due to the autoregressive nature of clip-by-clip synthesis, leading to noticeable error accumulation over time.

Another reason of error accumulation is the shortage of long-term information. The generation of each clip only depends on the last clip. Once introducing errors, the errors will be amplified irreversibly, the model lacks of the mechanism to refine and correct the error. 

Attribute drift also stems from the shortage of long-term information. Due to occlusion and complicate storyline, fixed local frames cannot provide enough information. For instance, we want to generate a girl whose face is not complete in the last clip due to back to the lens. If we do not retrieval the earlier frames to acquire more information of this girl, the appearance of this girl will be drift.

Based on these analysis, we formulate the following questions: besieds local frames, how to introduce long-term information to prevent from error accumulation and attribute drift? Can we restrain the generation of errors in the high-frequency component?

\subsection{Long video generation model}
The training process is divided into two stages. In the first stage, we fine-tune the base diffusion model on large-scale short video data to obtain a video extension model, which generates temporally coherent future frames conditioned on a few initial frames and a textual prompt. In the second stage, we further fine-tune this model on long video data to introduce causal attention across clips, analogous to large language models (LLMs), and leverage key-value (KV) caching to incorporate long-term temporal information. This two-stage training strategy is motivated by two key considerations. First, short video data is abundantly available, enabling the effective training of a video extension model at scale. Second, training video extension models typically require fine-tuning all parameters, which is computationally infeasible for long video data due to the substantial GPU memory demands. By decoupling the training process, we ensure both scalability and long-range modeling capability.
\vspace{-2mm}
\paragraph{Video extension model}
Existing approaches~\cite{zhang2023i2vgen,blattmann2023stable,xing2024dynamicrafter,wan2025} extend text-to-video (T2V) models to image-to-video (I2V) models by channel-wise concatenation of the conditional latent representation with the noise latent. We adopt similar strategy. We introduce a few conditional frames from a short video to control video synthesizing. Specifically, we compress the condition frames $F_c\in \mathbb{R}^{C\times N_c \times H\times W}$ by VAE $\mathcal{E}$, where $N_c$ is the number of condition frames, and then pad it with zero-filled frames along the temporal axis to obtain condition latent $z_c\in\mathbb{R}^{c\times n\times h\times w}$, where n is the temporal dimension of noise video latent $z_t$. Additionally, we introduce a binary mask $M\in\{0,1\}^{1\times n\times h \times w}$ where $1$ indicates the preserved frames and $0$ denotes the frames to be generated. Finally, we concatenate $z_t$, $z_c$, and $M$ along channel. Since the input of the video extension model has more channels than T2V model ($2c+1$ v.s. $c$), an additional linear projection layer is employed, which is initialized with zero values. 

During training, we employ a multi-task learning to enhance the model’s generalization capability. In addition to conditioning on the first frames, we also incorporate variants such as conditioning on the last frames, a combination of the first and last frames, and randomly dropping condition frames with a 0.1 drop rate. Unlike prior I2V methods, we do not rely on external models, such as CLIP~\cite{radford2021learning}, to encode conditional frames beyond the VAE. We empirically find that this simple yet effective method is sufficient for generating temporally coherent frames.

\begin{figure}[t]
        \centering
        \includegraphics[width=1.\linewidth]{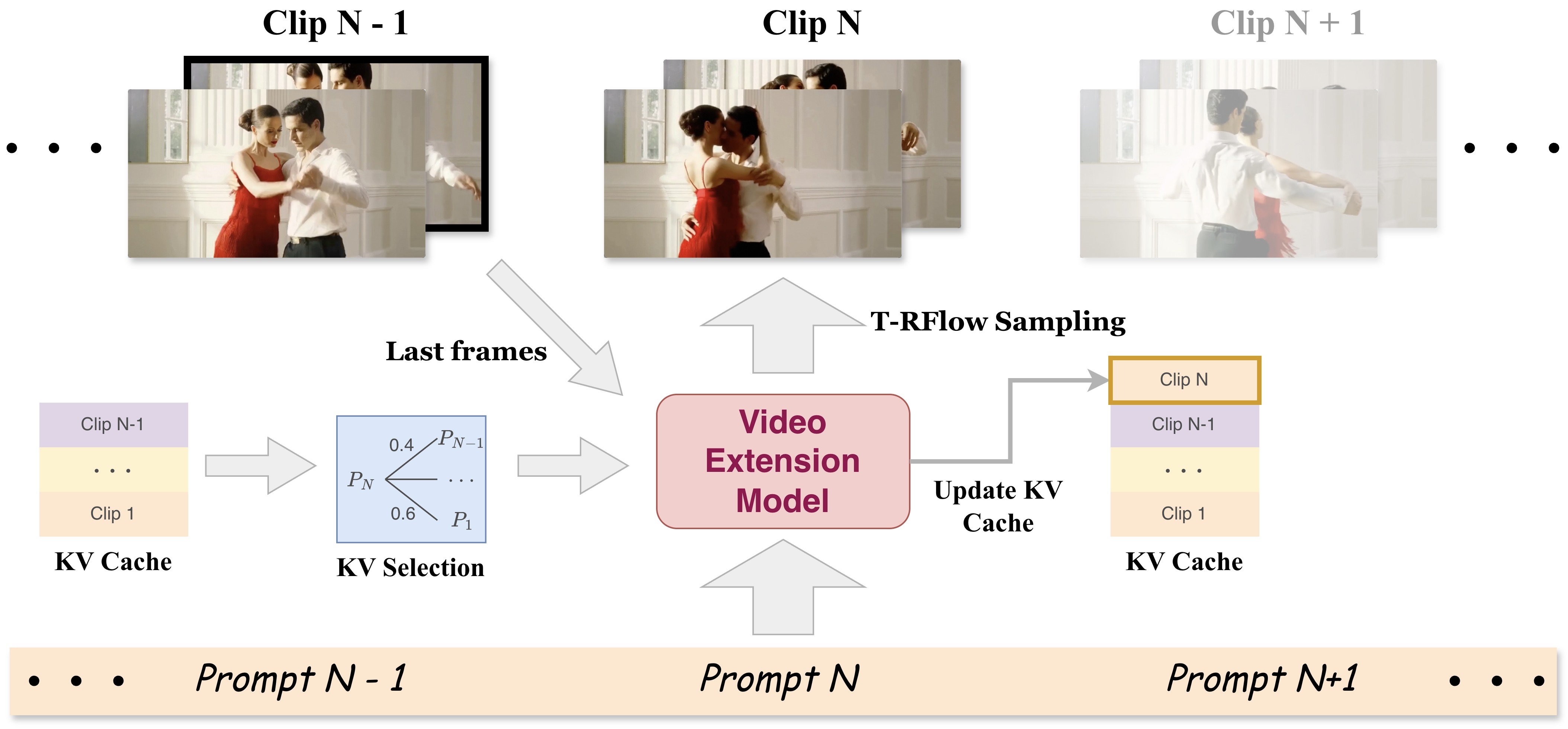}
\caption{Overview of our framework. The video extension model is from fine-tuning a video foundation model. After causal attention fine-tuning, this model generates long videos on a clip-by-clip paradigm, conditioned on KV cache and the last frames of the previous clip, thereby enhancing identity consistency and reducing error accumulation. To manage GPU memory efficiently, the KV cache is curated by our KV selection module. Additionally, the outputs at different timesteps are processed using our T-RFlow technique to further minimize error accumulation.  \vspace{-2mm}}
\label{fig:structure}
\end{figure}
\vspace{-2mm}
\paragraph{Causal Attention Fine-tuning}
To enable the model to learn long-term temporal dependencies, we fine-tune it on long videos, each of which is divided into $M$ short clips denoted by $[v^0, v^1, \cdots, v^M]$. In the original diffusion model, self-attention is computed within each clip. The query ($Q$), key ($K$), and value ($K$) of $i$-th clip $v^i$ are given by:
\begin{equation}
\label{eq4}
    Q_i=W_Qz_t^i,K_i=W_Kz_t^i,V_i=W_Vz_t^i,
\end{equation}
where $Q_i,K_i,V_i\in \mathbb{R}^{n\times h\times w\times c}$; $W_Q$, $W_K$, and $W_V$ are learnable projection weights. To enhance the ability to capture long-term information, we inject the KV from previous clips $v^{1:i-1}$ into the attention computation, and the $Q$, $K$, $V$ computation are reformulated as:
\begin{equation}
\label{eq5}
    Q_i=W_Qz_t^i,K_i=W_K[z_t^i,\hat{z}_0^{1:i-1}],V_i=W_V[z_t^i,\hat{z}_0^{1:i-1}], 
\end{equation}
where $Q_i\in \mathbb{R}^{n\times h\times w\times c}, K_i,V_i\in \mathbb{R}^{i\times n\times h\times w\times c}$; $\hat{z}_0^{1:i-1}$ is the set of latents of history clips from the last denoising step, which approximate the clean latent $z_0^{1:i-1}$. This design enables $KV$ reuse during inference, thus avoiding redundant computation.

Diffusion models commonly adopt Rotary Position Embedding (RoPE)~\cite{su2024roformer} in each Transformer block, which applies a rotary frequency matrix to token embeddings to encode relative positional relationships. We extrapolate RoPE along temporal dimension. With the clip-wise generation, the rotary angles of $Q$ and $K$ synchronously increase. This preserves intra-clip positional relationships while enabling extrapolation of inter-clip positions based on the temporal sequence.

However, applying $KV$ of all preceding clips becomes impractical as video length increases due to GPU memory limitations. To mitigate this, we select only a subset of historical $KV$ with the highest semantic similarity to the current clip, as measured by the similarity of their prompts using a LLM. When similarity scores are comparable, we prioritize earlier clips, since they tend to contain less error accumulation.

To further reduce memory usage and accelerate training, we employ Low-Rank Adaptation (LoRA)\cite{hu2022lora} for parameter-efficient fine-tuning. Inspired by StoryDiffusion\cite{zhou2024storydiffusion}, to prevent from overfitting, we inject $KV$ only into the latter half of the Transformer blocks. Consequently, LoRA is applied solely to the self-attention layers in those blocks, and only the LoRA parameters are updated during training.

\paragraph{Inference}
During inference, the video is generated clip-wisely. The generation of each clip is conditioned on individual prompt, last frames of last clip, and KV cache of previous clips. Similar to training, we employ a LLM or Sentence Bert~\cite{reimers2019sentence} to compute the prompt similarity to select KV.

\begin{figure*}[t]
        \centering
        \includegraphics[width=1.\linewidth]{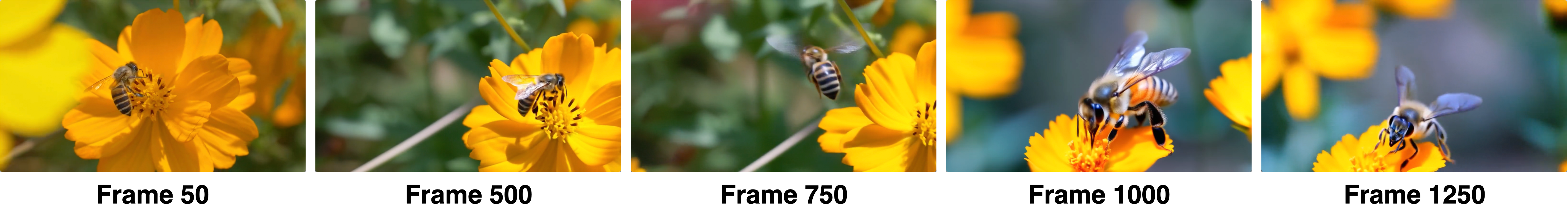}
\caption{A long video generated with T-RFlow. \vspace{-2mm}}
\label{fig:noerror}
\end{figure*}

\subsection{Truncation-rectified flow}
In Section~\ref{sec3.1}, we reformulate the final prediction as a linear combination of the predictions from all denoising steps. Our analysis reveals that the majority of error accumulation originates from the latter terms in Eq.\ref{eq3}, corresponding to the final denoising steps (as illustrated in Figure\ref{fig:error}). Owing to the linearity of Eq.~\ref{eq3}, each term can be individually manipulated during inference.
Interestingly, the middle terms are found to yield high-quality and stable predictions, while the late denoising steps tend to introduce visual artifacts. Based on above observation, we introduce a modified rectified flow, truncation-rectified flow (T-RFlow), to eliminate error accumulation in a training-free manner. Specially, we set a truncation threshold $T_1 \in (0, 1)$ and discard all terms corresponding to timesteps $t < T_1$ in order to suppress most of the accumulated degradation. However, removing early steps inevitably reduces the contribution of high-frequency components, which are critical for fine details.
To compensate for this, we introduce a second threshold $T_2 \in (T_1, 1)$ and remove all terms associated with timesteps $t > T_2$, which predominantly correspond to low-frequency structures. This truncation strategy re-balances the spectral content, increasing the relative contribution of high-frequency signals and enhancing overall visual fidelity.
$d_i$ in Eq~\ref{eq3} should be re-normalized, \ie, $d_i=di/\sum_{j\in [T_1, T_2]}d_j,i\in[T_1, T_2]$. 
To determine the value of $T_2$, we compute the value of average magnitude of the frequency spectrum (AF), which is defined as follows:
\begin{equation}
\label{eq6}
    AF = \frac{\sum_u\sum_vf(u,v)\cdot M(u,v)}{\sum_u\sum_vM(u,v)}, 
\end{equation}
where $M(u,v)$ represents the magnitude at frequency $(u,v)$, and $f(u,v)$ represents the radial frequency at $(u,v)$. We require that the AF of the video generated within the interval $[T_1, T_2]$ matches that of the video generated within the interval $[0, T]$.
We visualize the result in Figure~\ref{fig:noerror}. We can see that the error accumulation has been significantly suppressed and the detail of the close-up bee and the flower are retained well in the 4th sub-figure of Figure~\ref{fig:noerror}.

\begin{table*}[t]
  \centering
  \caption{\textbf{Comparison of different methods on VBench \cite{huang2024vbench}.} We report VBench metrics of different methods following the single-shot long video generation setting \cite{guo2025long,cai2025mixture}. Our method achieves superior performance in the majority of these metrics.\vspace{-3mm}}
  \label{tab:vbench}
  \setlength{\tabcolsep}{6pt}
  \renewcommand{\arraystretch}{1.2}
  \resizebox{\linewidth}{!}{
    \begin{tabular}{l|c|c|c|c|c|c}
      \toprule
      Method &
      \makecell{Subject \\ Consistency} $\uparrow$ &
      \makecell{Background \\ Consistency} $\uparrow$ &
      \makecell{Motion \\ Smoothness} $\uparrow$ &
      \makecell{Dynamic \\ Degree} $\uparrow$ &
      \makecell{Aesthetic \\ Quality} $\uparrow$ &
      \makecell{Image \\ Quality} $\uparrow$ \\
      \midrule
      MAGI-1~\cite{teng2025magi} & 0.8320 & 0.8931 & 0.9740 & 0.5537 & 0.5010 & 0.6120\\
      Self Forcing~\cite{huang2025self} & 0.8211 & 0.9050 & 0.9799 & 0.6015 & 0.5130 & 0.6218\\
      PAVDM~\cite{xie2025progressive} & 0.8415 & 0.9273 & 0.9769 & 0.6537 & 0.4970 & 0.6280 \\
      FramePack~\cite{zhang2025packing} & 0.9019 & 0.9450 & 0.9805 & 0.5715 & 0.5044 & 0.6381 \\
      SkyReels-V2-1.3B~\cite{chen2025skyreels} & 0.9391 & 0.9580 & 0.9838 & 0.6529 & 0.5320 & 0.6315 \\
      LCT~\cite{guo2025long} & 0.9380 & 0.9623 & 0.9816 & 0.6875 & 0.5200 & 0.6345 \\
      MoC~\cite{cai2025mixture} & 0.9398 & 0.9670 & 0.9851 & 0.7500 & \textbf{0.5547} & 0.6396 \\
      \midrule
      \textbf{Ours} & \textbf{0.9457} & \textbf{0.9691} & \textbf{0.9853} & \textbf{0.7569} & 0.5358 & \textbf{0.6620} \\
      \bottomrule
    \end{tabular}
  }
\vspace{-5mm}
\end{table*}

\section{Experiments}
\subsection{Implementation details}
We build upon the Wan2.1-T2V-1.3B model~\cite{wan2025}, which generates 5-second clips at 16 FPS with 480p resolution. Our method involves a two-stage training pipeline. In the first stage, we pretrain the Wan2.1 model, transforming it into a video extension model using data from our XunGuang-1.1M dataset. During this stage, we do not require long-form videos; instead, we segment them into clips ranging from 2 to 5 seconds. We randomly provide 0 to 9 initial frames and task the model with generating subsequent frames, thus enabling it to extend video content effectively. In the second stage, we perform causal attention fine-tuning using long-form videos from XunGuang-1.1M, supplemented by 10,000 private data entries. During fine-tuning, only the weights of the Keys and Values are optimized, enabling the model to generate coherent long video with multiple prompts. We retain up to 5 past clips' KV caches, selecting them based on prompt similarity computed by a sentence embedding model (specifically, Sentence-BERT in this work).

\vspace{-3mm}
\subsection{Main result}
We evaluate our method with state-of-the-art methods on VBench~\cite{huang2024vbench} under the single-shot long video generation setting~\cite{guo2025long,cai2025mixture}. As shown in Table~\ref{tab:vbench}, our method achieves superior performance across the majority of metrics, surpassing both open-source and large-scale proprietary baselines.
Specifically, our method obtains the highest scores in subject consistency (0.9457) and background consistency (0.9691), reflecting significant identity and attribute consistency in our method. motion smoothness (0.9870), dynamic degree (0.7720), aesthetic quality (0.5839). The highest image quality score (0.6620) shows the high-qualtiy of our video with minor error accumulation. Additionally, motion smoothness (0.9853) and dynamic degree (0.7569) surpass all the other methods. While MoC slightly outperforms our method in aesthetic quality (0.5547 v.s.\ 0.5538), our model delivers the most balanced overall performance across all six dimensions. These results highlight the effectiveness of our method in generating high-quality, dynamic, and consistent long video.

\begin{figure*}[t]
        \centering
        \includegraphics[width=1.\linewidth]{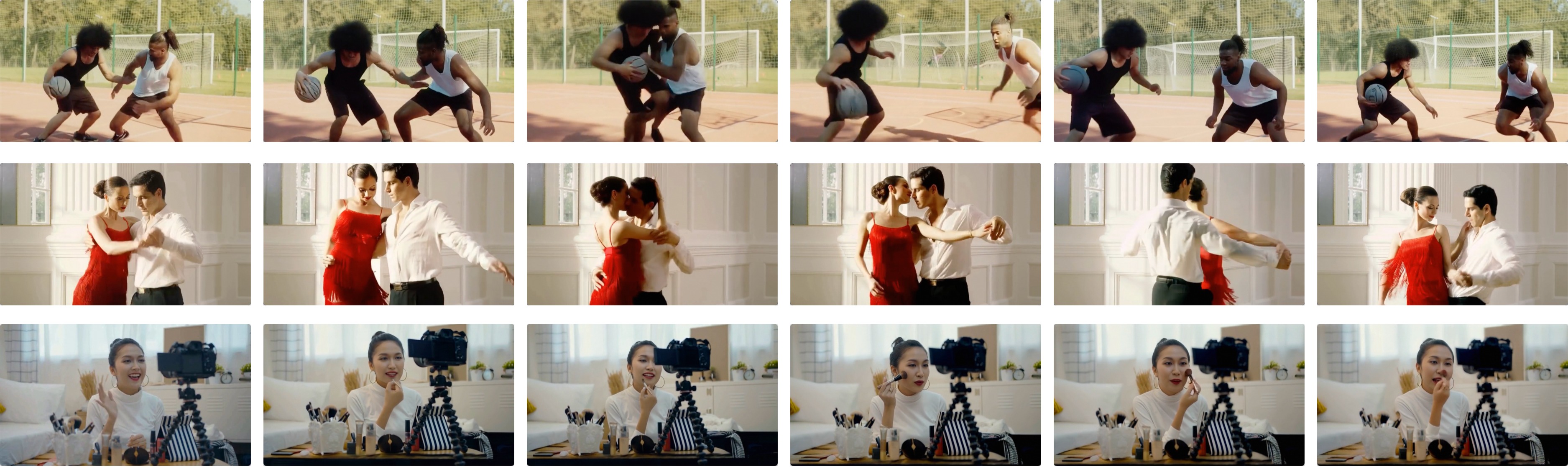}
\caption{Visualizations of our method. Each video has a 30-second duration and is generated using multiple prompts. The specific prompts used can be found in our supplementary material.}
\label{fig5}
\end{figure*}

The qualitative results in Figure~\ref{fig5}. Each video use multiple prompts to control the video content. Our approach, leveraging efficient KV caching and counteracting error accumulation, enables generated videos to depict characters making substantial movements while still preserving visual quality and coherence. This capability ensures that scenes remain free from accumulated errors over extended durations. For instance, in the first video, two individuals are vigorously playing basketball, exhibiting dynamic motion and transitioning between sunlight and shadow, yet maintaining consistent representation. In the second video, two dancers are able to execute complicate and dramatic actions, directed by our prompts, rather than simply swaying. Even when turning sideways or spinning, the consistency of the identities is preserved. These visual examples illustrate our method's effectiveness in enhancing both the dynamism and consistency of long videos while successfully mitigating error accumulation. 

\begin{table*}[t]
  \centering
  \caption{Performance evaluation of different module ablations. The baseline represents our complete model. We examine the effectiveness of discarding only the KV cache, as well as discarding both the KV cache and T-RF.\vspace{-3mm}}
  \label{tab2}
  \setlength{\tabcolsep}{6pt}
  \renewcommand{\arraystretch}{1.2}
  \resizebox{\linewidth}{!}{
    \begin{tabular}{l|c|c|c|c|c|c}
      \toprule
      Experiments &
      \makecell{Subject \\ Consistency} $\uparrow$ &
      \makecell{Background \\ Consistency} $\uparrow$ &
      \makecell{Motion \\ Smoothness} $\uparrow$ &
      \makecell{Dynamic \\ Degree} $\uparrow$ &
      \makecell{Aesthetic \\ Quality} $\uparrow$ &
      \makecell{Image \\ Quality} $\uparrow$ \\
      \midrule
      Baseline & 0.9457 & 0.9691 & 0.9853 & 0.7569 & 0.5358 & 0.6620 \\
      w/o KV Cache & 0.7811 & 0.8743 & 0.9850 & 0.7014 & 0.5271 & 0.6198 \\
      w/o KV cache and T-RF & 0.7046 & 0.8348 & 0.9640 & 0.6949 & 0.4803 & 0.5871 \\
      \bottomrule
    \end{tabular}
  }
\end{table*}

\subsection{Ablation study}
\paragraph{The effectiveness of KV cache finetuning and T-RFlow.}
We contend that our KV cache fine-tuning and T-RFlow effectively mitigate error accumulation. To assess their effectiveness, we conducted ablation experiments, the results of which are presented in Table~\ref{tab2}. As observed, omitting the KV cache fine-tuning results in decreased Vbench scores. These scores decline further when T-RFlow is not applied. This is visually demonstrated in Figure~\ref{fig6}. Without the previous KV cache, the video struggles to maintain color consistency, and without the use of T-RFlow, it suffers from significant error accumulation.

\begin{wrapfigure}{r}{0.4\textwidth}
  \centering
  \vspace{-25pt} 
  \captionof{table}{Image Quality scores under different truncation thresholds.} 
  \label{tab:wrap}
  \small
  \resizebox{0.9\linewidth}{!}{
  \begin{tabular}{@{}lccc@{}}
    \toprule
     & [0, 1] & [0, 0.7] & [0.3, 0.7] \\
    \midrule
    \makecell{Image \\ Quality} & 0.5871 & 0.6018 & 0.6198 \\
    \bottomrule
  \end{tabular}
  }
  \vspace{-15pt}
\end{wrapfigure}

\paragraph{The truncation thresholds of T-RFlow.}
To mitigate error accumulation, we introduce two truncation thresholds, $T_1$ and $T_2$, and experiment with different values to evaluate their effectiveness. Quantitative and qualitative results are presented in Table~\ref{tab:wrap} and Figure~\ref{fig7}, respectively. In the first sub-figure, where T-RFlow is not applied ($T_1=0$ and $T_2$=1),  we observe significant error accumulation in the video. By truncating the high-frequency component in RFlow and integrating over $t\in [0.3, 1]$, as shown in the second sub-figure, we achieve a notable reduction in error accumulation. However, this comes at the cost of reduced sharpness due to the loss of high-frequency details. To address this, we further remove early steps and integrate over $t\in [0.3, 0.7]$. This adjustment, illustrated in the third sub-figure, results in both lower error accumulation and improved video quality.

\paragraph{Error accumulation for longer videos.}
To verify that our method can significantly mitigate error accumulation, we assess image quality on 3-minute videos against a baseline that excludes both T-RFlow and KV cache mechanisms. As reported in Table~\ref{3min}, our approach successfully generates coherent long-form videos while significantly mitigating error accumulation, outperforming the baseline.

\begin{figure}[htbp]
    \centering
    \begin{minipage}[t]{0.48\linewidth}
        \centering
        \includegraphics[width=\linewidth]{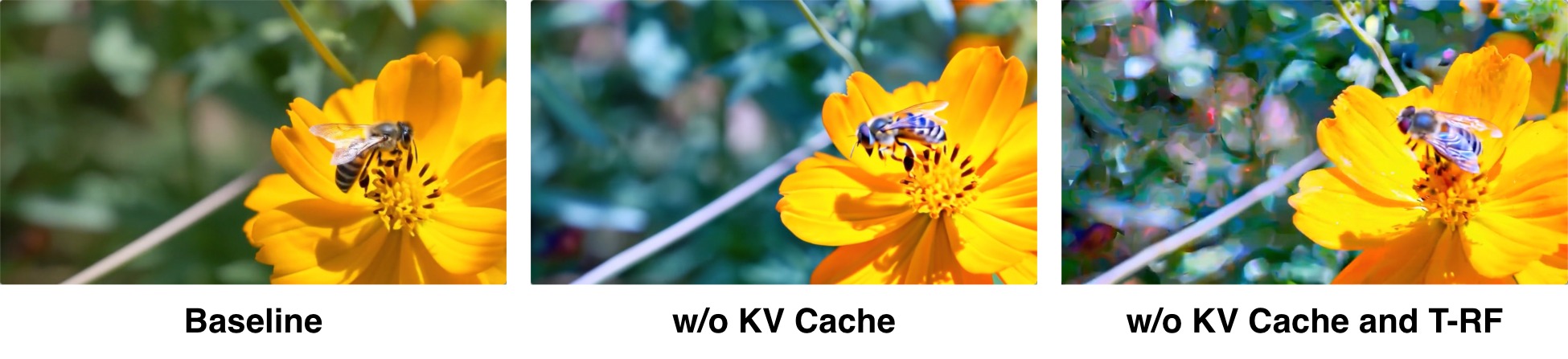}
        \caption{30-second videos under three different settings.}
        \label{fig6}
    \end{minipage}
    \hfill
    \begin{minipage}[t]{0.48\linewidth}
        \centering
        \includegraphics[width=\linewidth]{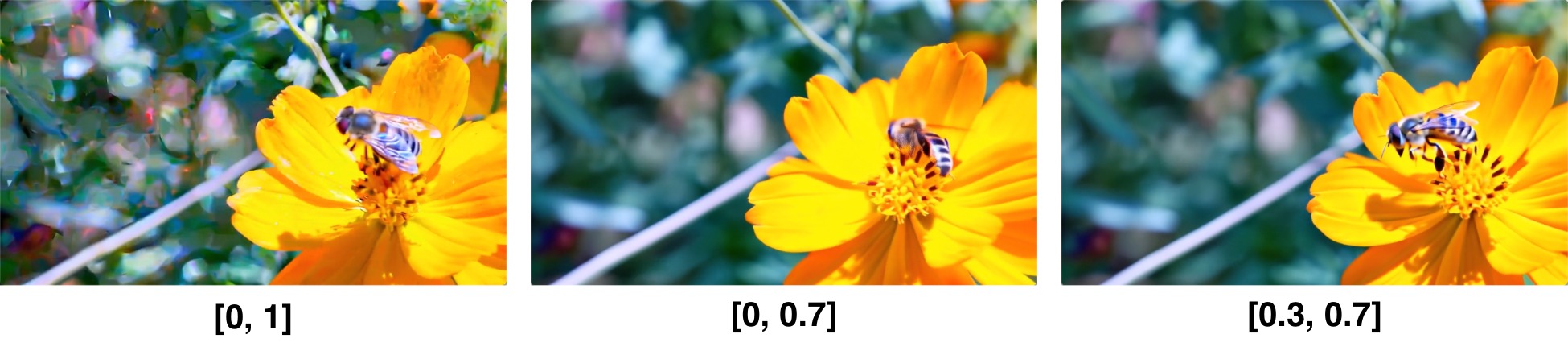}
        \caption{30-second videos under different truncation thresholds.}
        \label{fig7}
    \end{minipage}
\end{figure}

\begin{table}[t]
  \centering
  \caption{Image Quality of different video clips.}
  \label{3min}
  \setlength{\tabcolsep}{6pt}
  \renewcommand{\arraystretch}{1.2}
  \small
  \resizebox{\linewidth}{!}{
    \begin{tabular}{l|c|c|c|c|c|c}
      \toprule
      Experiments &
      0-30s &
      30-60s &
      60-90s &
      90-120s &
      120-150s &
      150-180s \\
      \midrule
      Ours & 0.6674 & 0.6600 & 0.6501 & 0.6414 & 0.6338 & 0.6140 \\
      Baseline & 0.6091 & 0.5489 & 0.4800 & 0.4049 & 0.3528 & 0.3075 \\
      \bottomrule
    \end{tabular}
  }
\end{table}

\section{Limitation}
In this work, our primary focus is on generating high-quality videos. However, our method currently does not support real-time video generation. In future work, we aim to explore strategies for streaming long video generation to address this limitation.

\section{Conclusion}

We present an novel framework for long video generation that addresses challenges inherent in long video synthesis, including error accumulation and attribute drift. By fine-tuning a diffusion model into a video extension model and utilizing causal attention across video clips, our method effectively leverages the power of large-scale short video datasets and adapts to long video sequences. The introduction of KV caching and truncation-rectified flow (T-RFlow) has proven pivotal in maintaining visual and identity consistency over extended durations. Moreover, our thorough experimentation demonstrates the superiority of the proposed method in video quality, and content coherence, setting new benchmarks for realistic minute-level video generation. 

%
%
\bibliographystyle{splncs04}
{\small

}
\end{document}